\documentclass{article}
\usepackage{ijcai26}

\usepackage{times}
\usepackage{soul}
\usepackage{url}
\usepackage[hidelinks]{hyperref}
\usepackage[utf8]{inputenc}
\usepackage[small]{caption}
\usepackage{graphicx}
\usepackage{amsmath}
\usepackage{amsfonts}
\usepackage{amsthm}
\usepackage{booktabs}
\usepackage{algorithm}
\usepackage{algorithmic}
\usepackage[switch]{lineno}
\usepackage{multirow}
\usepackage{makecell}
\usepackage{tabularx} 
\usepackage{xcolor}
\usepackage{array}
\usepackage{amssymb} 
\usepackage{makecell}

\title{Embedding-Aware Feature Discovery: Bridging Latent Representations and Interpretable Features in Event Sequences}

\title{Embedding-Aware Feature Discovery: Bridging Latent Representations and Interpretable Features in Event Sequences}

\author{
	Artem Sakhno$^1$ \and
	Ivan Sergeev$^1$ \and
	Alexey Shestov$^1$ \and
	Omar Zoloev$^1$ \and
	Elizaveta Kovtun$^1$ \and
	Gleb Gusev$^1$ 
	Andrey Savchenko$^{1,2}$ \and
	Maksim Makarenko$^1$
	\affiliations
	$^1$ Sber AI Lab\\
	$^2$ ISP RAS Research Center for Trusted AI
}

\begin{document}
\maketitle

\begin{abstract}
Industrial financial systems operate on temporal event sequences such as transactions, user actions, and system logs. While recent research emphasizes representation learning and large language models, production systems continue to rely heavily on handcrafted statistical features due to their interpretability, robustness under limited supervision, and strict latency constraints. This creates a persistent disconnect between learned embeddings and feature-based pipelines.
We introduce \emph{Embedding-Aware Feature Discovery} (EAFD), a unified framework that bridges this gap by coupling pretrained event-sequence embeddings with a self-reflective LLM-driven feature generation agent. EAFD iteratively discovers, evaluates, and refines features directly from raw event sequences using two complementary criteria: \emph{alignment}, which explains information already encoded in embeddings, and \emph{complementarity}, which identifies predictive signals missing from them.
Across both open-source and industrial transaction benchmarks, EAFD consistently outperforms embedding-only and feature-based baselines, achieving relative gains of up to $+5.8\%$ over state-of-the-art pretrained embeddings, resulting in new state-of-the-art performance across event-sequence datasets.
\end{abstract}

\section{Introduction}

Industrial machine learning systems commonly operate on temporal event sequences, including financial transactions~\cite{mollaev2025multimodal,latte}, online purchase histories~\cite{liu2025enhancing}, electronic health records~\cite{draxler2025transformers}, spatial trajectories~\cite{unitraj2025}. In production environments, these sequences are processed under strict constraints on latency, throughput, and stability~\cite{xiao2024deep}, which has led to the widespread adoption of specialized encoder models that produce compact embeddings~\cite{brisaboa2018compact}, alongside handcrafted statistical interpretable features that encode domain-specific structure~\cite{verdonck2024special,topology}. While recent research has increasingly focused on representation learning~\cite{shestov2025llm4es}, real-world systems continue to rely heavily on feature-based pipelines~\cite{louhi2023empirical,faubel2024mlops} due to their ease of configuration for new tasks with limited labeled data, interpretability, and predictable runtime behavior.

\begin{figure}[t]
\centering
\includegraphics[width=0.99\linewidth]{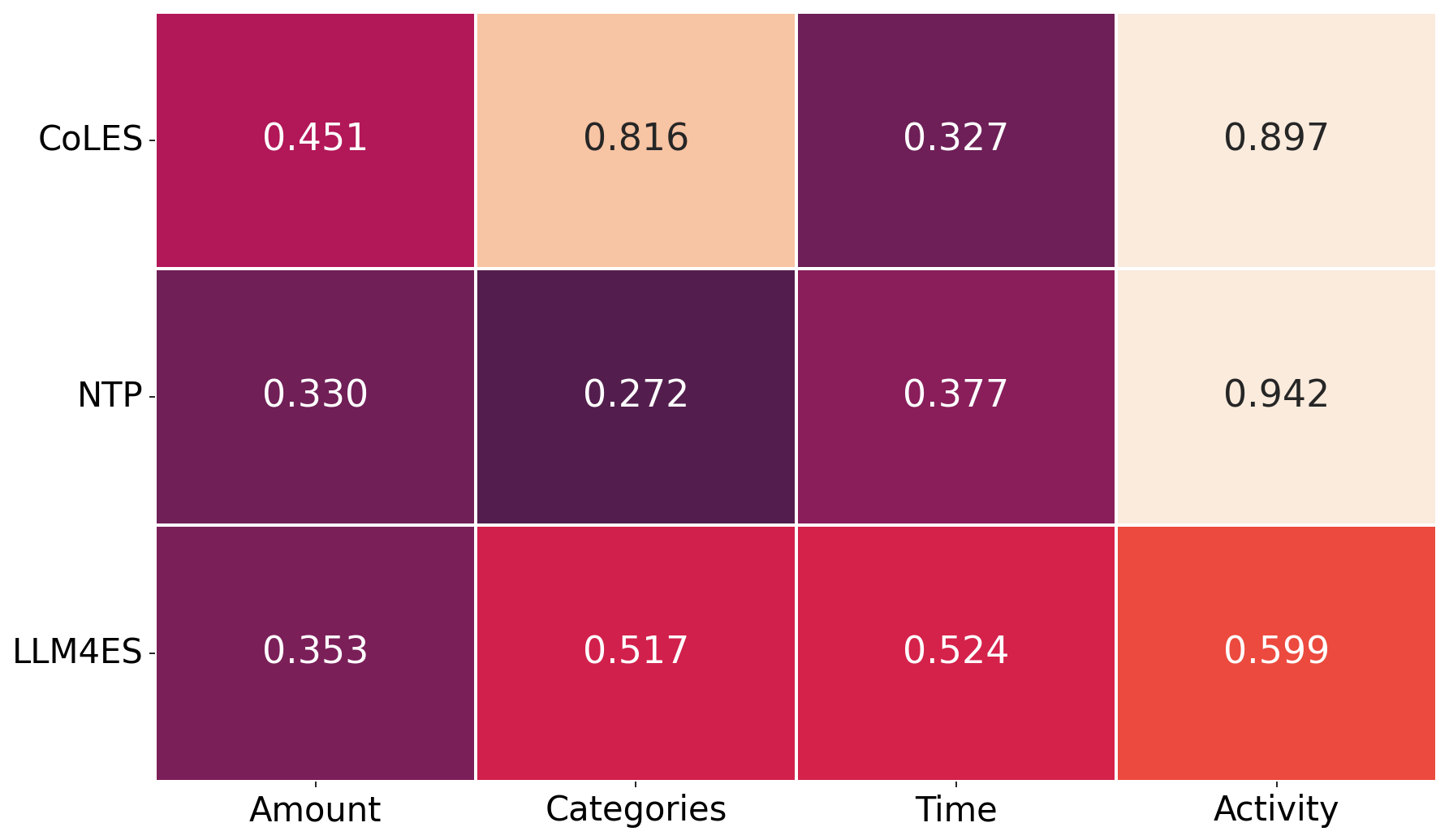}
\caption{\textbf{Blind spots of embeddings.} Coefficient of determination ($R^2$) of a feature reconstruction based on embedding representations CoLES, NTP, and LLM4ES on the Rosbank dataset. Values highlight systematic representational blind spots across embeddings.} 
\label{fig}
\end{figure}

Recent advances in automated feature engineering ~\cite{de2025automated} and AutoML~\cite{gijsbers2024amlb} have greatly reduced the manual effort required to construct and select features for structured data. Many end-to-end AutoML frameworks incorporate feature selection directly into the training pipeline~\cite{arik2021tabnet}, while other approaches focus explicitly on automated feature construction. For example, AutoFeat \cite{horn2019autofeat} generates non-linear transformations for tabular data, and Deep Feature Synthesis (DFS), implemented in Featuretools \cite{kanter2015deep}, produces fixed-size representations from sequential inputs using predefined aggregation operators. More recently, research in automated feature engineering has drifted toward the use of large language models (LLMs) as reasoning and orchestration components, moving toward fully automated and increasingly agentic frameworks that coordinate feature generation, selection, and evaluation \cite{hollmann2023caafe,guo2024ds,abhyankar2025llm}. Despite this progress, automated feature engineering for sequential data largely remains decoupled from representation learning.

As a result, deep embeddings and statistical features are optimized through largely separate workflows. The consequences of this separation become apparent when examining how well embeddings reconstruct different classes of handcrafted features (Figure~\ref{fig}). While embeddings capture certain aggregate statistics with high fidelity, other feature groups exhibit persistently low reconstruction quality, revealing systematic representational blind spots that vary across embedding models. This is problematic in practice: information missing from the embedding cannot be recovered by subsequent models, leading to performance ceilings and redundant feature construction. In the absence of a clear interface between representation learning and feature discovery, feature engineering remains unaware of representation gaps, and embedding refinement proceeds without interpretable guidance from structured signals. 
To address this gap:
\begin{itemize}
\item We introduce \textit{Embedding-Aware Feature Discovery} (EAFD)\footnote{The source code is provided in the supplementary materials.}, a unified framework that transforms the disconnect between embeddings and structured features into an iterative, self-reflective reasoning process for feature discovery (Section~\ref{sec:method}). EAFD operates in two complementary regimes: an \emph{interpretability regime}, which analyzes and explains the information already encoded in pretrained embeddings by aligning with interpretable features, and a \emph{performance regime}, which discovers complementary features that provide additional predictive signal not present in embeddings. EAFD anchors on pretrained embeddings and integrates them with an LLM-driven feature-generation agent that actively explores raw event-sequence data, proposing and evaluating candidate features based on alignment and complementarity criteria.

\item We evaluate EAFD on four open-source event-sequence benchmarks, where it consistently outperforms embedding-only and feature-based baselines across datasets and backbone embeddings (Section~\ref{sec:performance}). EAFD achieves relative gains of up to $+5.8\%$ over state-of-the-art embeddings and up to $+19\%$ over weaker representations, establishing a new state of the art on open-source transaction datasets. We further validate EAFD on a large-scale, proprietary, multitarget financial dataset, where a single EAFD-enhanced representation improves across classification and regression targets, yielding gains of up to $ 12.55\%$ and up to $3.87\%$ error reduction, respectively.

\item We use EAFD's explicit feedback to analyze multiple state-of-the-art embedding models, uncovering systematic representational biases and information gaps (Section~\ref{sec:interpret}). Leveraging these insights, we propose several modifications for the CoLES framework, resulting in up to 1.20\% relative improvement in churn prediction. We also demonstrate how EAFD’s embedding interpretability signals can be further coupled with a privacy-preserving feature erasure mechanism~\cite{seputis2025rethinking,liu2025urania} that identifies and suppresses sensitive attributes encoded in embeddings.

\end{itemize}

Overall, EAFD provides a unified and practical framework for analyzing and enhancing event-sequence embeddings through feature discovery, bridging representation learning and feature-based pipelines across both benchmark datasets and large-scale industrial systems.

\begin{figure*}[h!]
  \centering
  \includegraphics[width=\textwidth]{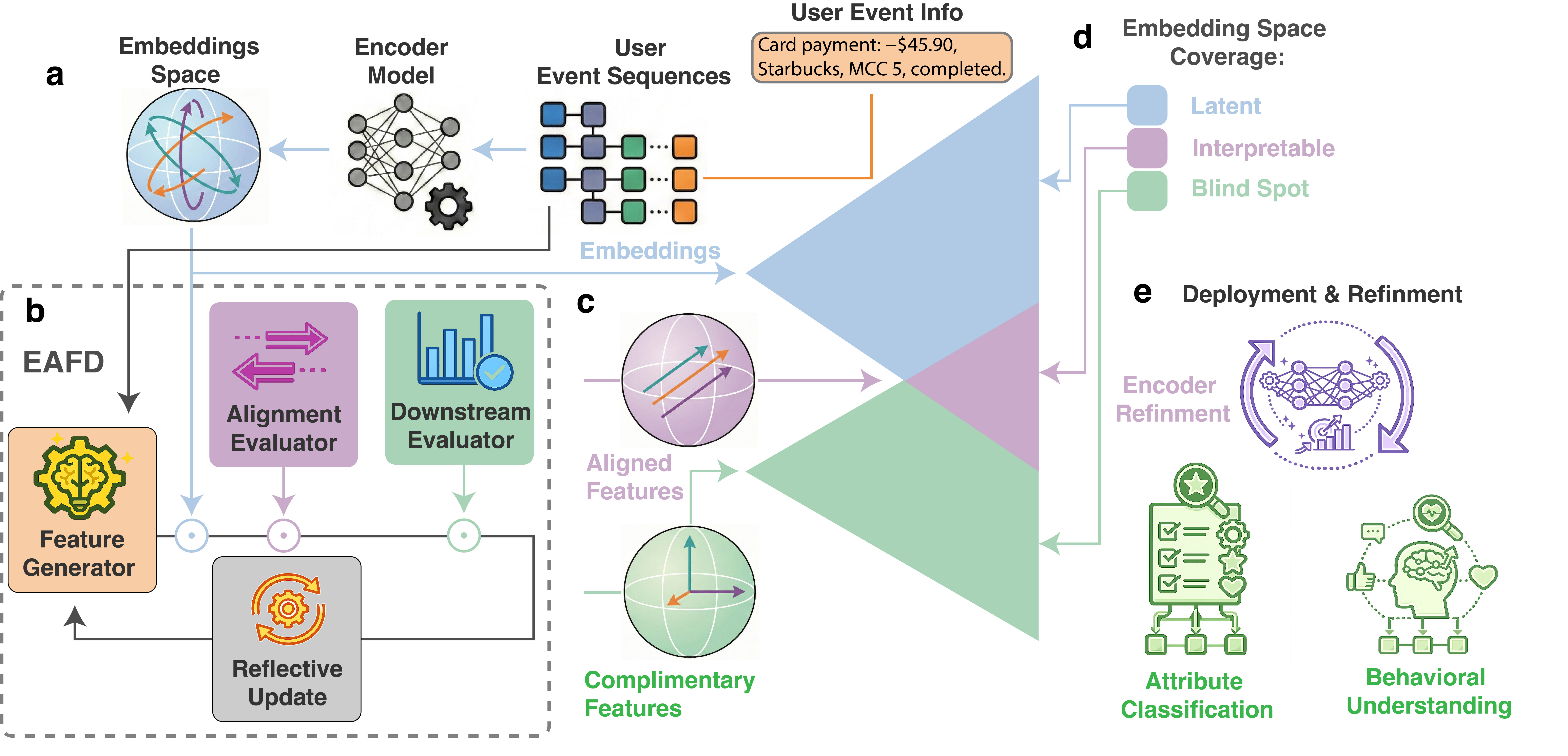}
\caption{
\textbf{Overview of Embedding-Aware Feature Discovery (EAFD).}
\textbf{(a)} Latent embedding pipeline mapping event sequences to continuous user representations.
\textbf{(b)} EAFD agent loop: an LLM-based generator proposes interpretable features from raw sequences, evaluated by embedding–feature alignment and downstream utility, with reflective updates guiding iteration.
\textbf{(c)} Feature outcomes: \emph{aligned} features recover information encoded in the embedding, while \emph{complementary} features capture predictive factors missing from it.
\textbf{(d)} Embedding–feature space decomposition into latent, interpretable, and blind-spot regions.
\textbf{(e)} Deployment and refinement: discovered features improve downstream tasks and support targeted encoder refinement (e.g., coverage, robustness, privacy).
}

  \label{fig:eafd_overview}
\end{figure*}

\section{Related Works}


Automated feature engineering ({\it AutoFE})~\cite{de2025automated} has been widely studied for tabular data. Classical systems range from LightAutoML's data cleaning~\cite{vakhrushev2021lightautoml} to more complex approaches like Featuretools~\cite{kanter2015deep} and AutoCross~\cite{li2024autocross}, which use predefined transformation grammars and cross-feature search. Neural methods, such as TabNet~\cite{arik2021tabnet}, learn latent feature representations directly, while OpenFE~\cite{zhang2023openfe} improves generation diversity and efficiency via a fast two-stage pruning.

Recent progress in large language models has extended this line of work toward feature-generation agents. CAAFE~\cite{hollmann2023caafe} shows that LLMs can synthesize meaningful features and executable code from natural-language descriptions, while follow-up approaches explore evolutionary search (LLM-FE~\cite{abhyankar2025llm}), program synthesis (DS-Agent~\cite{guo2024ds}), and table-structure reasoning~\cite{han2025tabular}. Both classical AutoFE systems and LLM-based feature agents exhibit a common limitation in modern industrial pipelines that rely on pretrained encoders: feature construction is performed independently of the embedding space, without assessing redundancy or complementarity, and without explicit reasoning over temporal structure in event-sequence data.

A complementary research direction investigates embedding probing~\cite{koprobing} and representation analysis~\cite{tinaz2025emergence}. Techniques such as linear probes~\cite{basile2025head}, sparse decomposition~\cite{harle2025measuring}, and centered kernel alignment~\cite{Maniparambil_2024_CVPR} provide insight into what pretrained models encode. These approaches have been applied extensively in language and vision models~\cite{li2025does} and, more recently, in representation learning for tabular and sequential data~\cite{tennenholtzdemystifying}. 
However, probing methods are diagnostic only: they reveal alignment or separability properties of embeddings but do not generate new features or provide actionable guidance for downstream modeling.

Event-sequence modeling further amplifies structural gap between feature construction and representation analysis, as temporal patterns such as recency, burstiness, and seasonality typically require explicit feature construction alongside specialized sequence encoders~\cite{klenitskiy2025encode}. Existing AutoFE and LLM-based feature agents do not incorporate feedback from pretrained event-sequence embeddings, either generating features without accounting for what the embedding already encodes~\cite{kanter2015deep} or analyzing embeddings without producing complementary symbolic features~\cite{hollmann2023caafe,abhyankar2025llm}. As a result, the problem of joint, embedding-aware feature discovery remains largely unaddressed.

\section{Proposed Method}
\label{sec:method}

Embedding-Aware Feature Discovery (EAFD) is an iterative framework for discovering interpretable and complementary features for event-sequence embeddings. Figure~\ref{fig:eafd_overview} illustrates the overall workflow. Given a collection of event sequences $\mathcal{S} = \{s_i\}_{i=1}^N$ with downstream labels $y_i$, a pretrained encoder model $f_\theta$ maps each sequence to a fixed-dimensional embedding $\mathbf{z}_i = f_\theta(s_i)$ (Figure~\ref{fig:eafd_overview}a). The encoder parameters $\theta$ are frozen throughout feature discovery and serve as a stable representation anchor. Feature candidates are defined as deterministic and interpretable mappings $g_k : \mathcal{S} \rightarrow \mathbb{R}^M$, and are proposed by an LLM-based generator $G$ conditioned on the raw event data and accumulated reflective feedback from previous iterations (Figure~\ref{fig:eafd_overview}b).

To guide the discovery process, EAFD operates in two compatible regimes: interpretability and performance. 

\paragraph{Interpretability Regime.}This regime aims to explain the information encoded in the latent space by aligning it with human-interpretable concepts. Here, we evaluate whether a generated feature $g_k$ recovers information already present in $\mathbf{z}$. This is quantified by the \emph{Alignment Score}, which measures the predictive consistency or correlation between the embeddings and the feature-based predictor:\begin{equation}A(g_k) = \psi\bigl(\mathbf{z}, g_k(\mathcal{S})\bigr),\end{equation}where $\psi$ is the alignment metric ($R^2$ of a small gradient boosting model mapping generated features to embedding vectors).
High $A(g_k)$ indicates that the feature effectively translates latent dimensions into interpretable logic.

\paragraph{Performance Regime.}This regime seeks to supplement the embeddings with additional predictive signals. The goal is to identify ``blind spots'', information relevant to the task but missing from $\mathbf{z}$. We measure this via the \emph{Downstream Utility Score}, which compares the loss of an embedding-only model with that of a joint model utilizing both representations:\begin{equation}U(g_k) = \mathcal{L}\bigl(\mathbf{z}, y\bigr) - \mathcal{L}\bigl([\mathbf{z}, g_k(\mathcal{S})], y\bigr).\end{equation}A positive $U(g_k)$ implies that the feature captures complementary structure that improves downstream performance.

While distinct in their objectives, these two regimes are employed simultaneously to drive the iterative feedback loop. To introduce additional features, the generator $G$ is conditioned on a detailed reflection from the previous step. In the interpretability regime, this reflection includes the alignment scores $A(g_k)$ alongside qualitative examples that distinguish features strongly coupled with $\mathbf{z}$ from those that are orthogonal. In the performance regime, the feedback comprises the utility scores $U(g_k)$ and explicit feature importance rankings derived from the downstream model on the validation set. 

The generator translates these reflective insights into executable code definitions for new candidate features. To ensure robustness, if the generated code fails to execute (e.g., due to syntax errors or runtime exceptions), the framework triggers a self-correction \emph{debug mode} to fix the implementation iteratively. Successfully executed features are then evaluated and, based on the aggregated signals shown in Figure~\ref{fig:eafd_overview}c, categorized as \emph{aligned} ($A(g_k)$ high, $U(g_k)\!\approx\!0$) or \emph{complementary} ($U(g_k)\!>\!0$). The categorization induces a decomposition of information across the joint feature-embedding space (Figure~\ref{fig:eafd_overview}d): some components remain purely latent in the embedding, others become interpretable through alignment with discovered features, and blind spots correspond to feature dimensions that supplement the embedding with missing task-relevant information. 

The resulting features can be used directly in downstream tasks (Figure~\ref{fig:eafd_overview}e), such as attribute classification and behavioral understanding (Section \ref{sec:performance}). The identified embedding blind spots can guide targeted encoder refinement, such as controlled adaptation or selective information erasure (Section \ref{par:privacy}), for privacy-sensitive applications.

\begin{table*}[t]
\centering
\small
\renewcommand{\arraystretch}{1.25}

\begin{tabularx}{\textwidth}{|X|cc|cc|cc|cc|}
\hline
Method 
& \multicolumn{2}{c|}{Age} 
& \multicolumn{2}{c|}{Gender} 
& \multicolumn{2}{c|}{Rosbank} 
& \multicolumn{2}{c|}{DataFusion} \\

& Acc & $\Delta$\% 
& AUC & $\Delta$\% 
& AUC & $\Delta$\% 
& AUC & $\Delta$\% \\
\hline

CoLES 
& 0.645 & -- 
& 0.888 & -- 
& 0.835 & -- 
& 0.738 & -- \\
\hline

NTP 
& 0.540 & -- 
& 0.849 & -- 
& 0.798 & -- 
& 0.648 & -- \\
\hline

LLM4ES 
& 0.651 & -- 
& -- & -- 
& 0.849 & -- 
& -- & -- \\
\hline\hline

Agg Features + CoLES
& 0.649 & \textcolor{green}{+0.62} 
& 0.889 & \textcolor{green}{+0.11} 
& 0.832 & \textcolor{red}{-0.36} 
& 0.740 & \textcolor{green}{+0.27} \\
\hline

Featuretools + CoLES
& 0.644 & \textcolor{red}{-0.16}  
& 0.889 & \textcolor{green}{+0.11}  
& 0.863 & \textcolor{green}{+3.35} 
& 0.760 & \textcolor{green}{+2.98} \\
\hline

LLMFE (Agg Features + CoLES) 
& 0.644 & \textcolor{red}{-0.16} 
& 0.887 & \textcolor{red}{-0.11} 
& 0.825 & \textcolor{red}{-1.20} 
& 0.740 & \textcolor{green}{+0.27} \\
\hline

LLMFE (Featuretools + CoLES) 
& 0.644 & \textcolor{red}{-0.16} 
& 0.888 & 0.00 
& 0.862 & \textcolor{green}{+3.23} 
& 0.760 & \textcolor{green}{+2.98} \\
\hline

CAAFE (Agg Features + CoLES) 
& 0.638 & \textcolor{red}{-1.09}  
& 0.877 & \textcolor{red}{-1.24} 
& 0.840 & \textcolor{green}{+0.60} 
& 0.747 & \textcolor{green}{+1.22} \\
\hline

CAAFE (Featuretools + CoLES) 
& 0.641 & \textcolor{red}{-0.62}  
& 0.882 & \textcolor{red}{-0.68}  
& 0.870 & \textcolor{green}{+4.19} 
& 0.762 & \textcolor{green}{+3.25} \\

\hline \hline

EAFD (CoLES) 
& 0.649 & \textcolor{green}{+0.62} 
& \textbf{0.898} & \textcolor{green}{\textbf{+1.13}} 
& \textbf{0.872} & \textcolor{green}{\textbf{+4.43}} 
& \textbf{0.781} & \textcolor{green}{\textbf{+5.83}} \\
\hline

EAFD (NTP) 
& 0.623 & \textcolor{green}{+15.37} 
& 0.870 & \textcolor{green}{+2.47} 
& 0.865 & \textcolor{green}{+8.40} 
& 0.773 & \textcolor{green}{+19.29} \\
\hline 

EAFD (LLM4ES) 
& \textbf{0.652} & \textcolor{green}{+0.15} 
& -- & -- 
& 0.866 & \textcolor{green}{+2.00} 
& -- & -- \\
\hline
\end{tabularx}
\caption{Performance comparison of EAFD against baseline embeddings and feature-based agents. Relative improvement ($\Delta$\%) is reported w.r.t. the corresponding backbone embedding. Standard deviations are approximately: Age $\pm0.004$, Gender $\pm0.007$, Rosbank $\pm0.005$, DataFusion $\pm0.008$.}
\label{tab:public_results}
\end{table*}

\section{Experiments}

This section evaluates the framework in two parts. The first part focuses on downstream performance when complementary features are iteratively added to pretrained embeddings. The second part examines interpretability and analyzes how feature based projections reveal the structure of the embedding space.

\subsection{Experimental Settings}
\subsubsection{Datasets}
All datasets consist of transactional or behavioral event sequences. We evaluate EAFD on four public financial benchmarks and one large-scale proprietary industrial dataset. \\
\textbf{Age Prediction}~\cite{sberbank_sirius} contains $\sim$44M transactions from 50K users, with relative timestamps, transaction categories, and anonymized amounts. Following a semi-supervised setup, 20K unlabeled sequences are used for embedding pretraining, while 30K labeled sequences are used for EAFD training and downstream evaluation.\\
\textbf{Gender Prediction}~\cite{sberbank_gender} is a demographic benchmark with 8.4K labeled user sequences, each defined by relative timestamps, merchant category codes, and transaction types, used to assess projection consistency across targets.\\
\textbf{Rosbank}~\cite{rosbank1} includes transaction histories for 10K users over a three-month period, with churn labels available for only 5K users. The remaining unlabeled sequences are leveraged for self-supervised embedding learning, enabling evaluation in a label-constrained regime.\\
\textbf{DataFusion}~\cite{DataFusion2024} targets churn prediction over a six-month horizon using nine months of transactional history. The dataset comprises 13M transactions from 96K users (64K labeled), with each record including timestamps, merchant category codes, currency, and amount.\\
\textbf{Proprietary Multi-target Dataset.} We further validate EAFD on a private banking dataset with millions of users, requiring simultaneous prediction of age group, gender, and a continuous financial outcome forecasted one year ahead, reflecting a realistic industrial multi-target setting.

\subsubsection{Implementation Details}

We use gpt-oss-120b via vLLM on 4$\times$NVIDIA A100 GPUs. We set the context window to 100K tokens and the maximum output length to 16K tokens. We run EAFD for 5 iterations and evaluate the generated features using a CatBoost~\cite{prokhorenkova2018catboost} model.

\subsection{Performance Evaluation}
\label{sec:performance}

We begin with experiments that use embeddings produced by a single encoder model. For each dataset, a pretrained encoder $f_\theta$ generates user embeddings $z_i = f_\theta(s_i)$ of fixed dimensionality. To evaluate method performance with embeddings that capture distinct event-sequence information, we use three self-supervised encoders: CoLES~\cite{babaev2022coles} (RNN, contrastive loss), NTP~\cite{padhi2021tabular} (RNN, NTP loss), and LLM4ES~\cite{shestov2025llm4es} (fine-tuned LLM, text‑enriched sequences, NTP loss). The framework is run with a fixed feature budget, and the complementary features with the highest uplift are selected.

We compare the following approaches:

\begin{itemize}
\item \textbf{Embedding-only}: $z_i$ used as the sole input to the downstream model.
\item \textbf{Aggregation features}: statistical event-sequence aggregations from pytorch-lifestream~\cite{sakhno2025pytorch} and Featuretools~\cite{kanter2015deep}, concatenated with $z_i$.
\item \textbf{LLM-based feature agents}: CAAFE and LLMFE, which generate high-level features from tabular inputs composed of baseline aggregations and embeddings.
\item \textbf{Non-LLM feature generation}: OpenFE~\cite{zhang2023openfe}, generating features from tabular inputs without joint model training.
\item \textbf{AutoML pipelines}: LightAutoML~\cite{vakhrushev2021lightautoml}, performing integrated feature engineering and model selection on aggregated representations.
\item \textbf{EAFD (ours)}: EAFD-generated features combined with $z_i$ under the same feature budget.
\end{itemize}

Table~\ref{tab:public_results} reports results on open-source financial datasets for different backbone embeddings. Across all datasets and tasks, EAFD consistently improves performance over the corresponding base representations, demonstrating that the discovered features capture complementary information not fully encoded in the embedding space. The most pronounced gains occur for NTP embeddings, where EAFD yields large relative improvements across all datasets, including +15.4\% on Age Prediction and up to +19.3\% on the DataFusion benchmark. EAFD effectively recovers task-relevant structure that is weakly represented in NTP embeddings. EAFD also improves performance for stronger backbones: when applied to CoLES, it achieves consistent gains across all datasets, with particularly large improvements on Rosbank and DataFusion, while EAFD applied to LLM4ES yields smaller but stable improvements. 

\subsubsection{Iterative Feature Generation Dynamics}

\begin{figure}[h]
    \centering
    \includegraphics[width=0.95\linewidth]{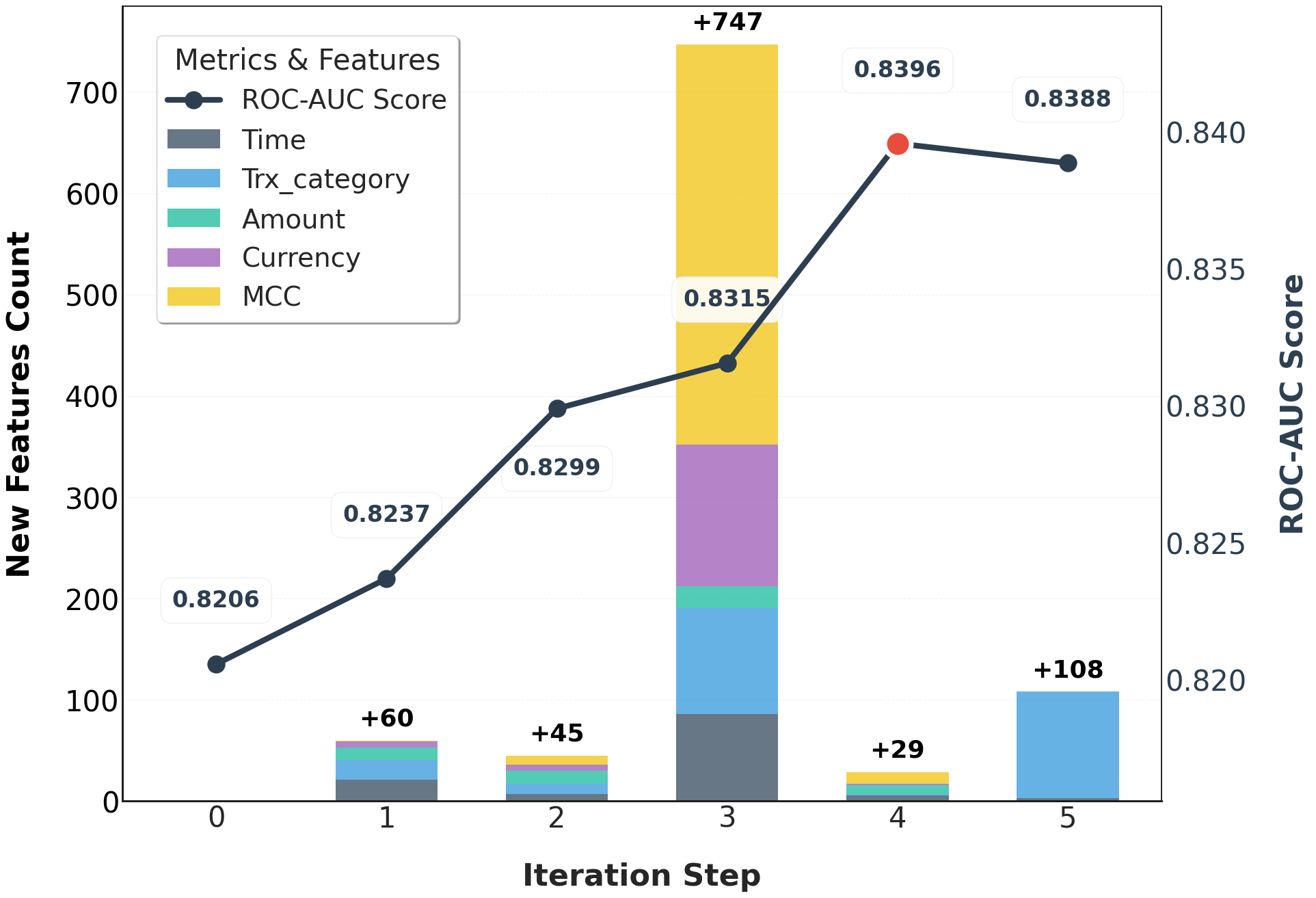}
    \caption{\textbf{EAFD iteration dynamics} of Rosbank validation ROC-AUC and feature composition across EAFD iterations.}
    \label{fig:iteration_dynamics}
\end{figure}

Figure~\ref{fig:iteration_dynamics} illustrates the iterative nature of the EAFD framework and its impact on the Rosbank prediction task. The line plot tracks the steady increase in the validation ROC-AUC score, while the stacked bars detail the volume and functional categories of features added at each step. We observe a consistent improvement in performance, rising from a baseline of $0.8206$ to a peak of $0.8396$ by the fourth iteration.

In total, EAFD utilizes 37 distinct aggregation functions over the five iterations. Qualitative analysis reveals that the most significant metric gains are often driven by interpretable temporal features, such as the time elapsed between the first and last transaction or transaction frequency within recent windows (e.g., count in the last 30 days). However, the framework also successfully identifies complex, high-utility signals that capture behavioral nuances, including the Herfindahl–Hirschman Index (HHI) on MCC distributions, Exponentially Weighted Moving Averages (EWMA), and autocorrelation features applied to transaction amounts.

\subsubsection{Multi-target Feature Generation}
\label{sec:multitarget}
In this section, we validate EAFD on a large-scale industrial multi-target dataset. Table~\ref{tab:private_results} summarizes the experimental results, which demonstrate two critical capabilities of the proposed framework. 

First, EAFD consistently improves the performance of state-of-the-art pretrained embeddings. When applied to strong baselines such as $\text{CoLES}$ and $\text{NTP}$, EAFD reduces regression Mean Absolute Error (MAE) by $3.87\%$ and $1.55\%$, respectively. These results show that pretrained latent representations do not fully capture all task-relevant information, and that EAFD systematically recovers additional predictive signal through complementary feature discovery.

\begin{table}[h]
\centering
\small
\begin{tabular}{|l|c|c|c|}
\hline
\textbf{Method} & \multicolumn{3}{c|}{\textbf{Private Dataset}} \\ \cline{2-4}
 & \textit{Age} & \textit{Gender} & \textit{Regression} \\ \hline
CoLES & 0.743 & 0.898 & 11373 \\ \hline
NTP & 0.736 & 0.895 & 10839 \\ \hline
LLM4ES & 0.692 & 0.789 & 11462 \\ \hline

\textbf{EAFD (CoLES)} 
& \makecell{\textbf{0.756} \\ \textcolor{green}{(+1.75\%)}} 
& \makecell{\textbf{0.901} \\ \textcolor{green}{(+0.33\%)}} 
& \makecell{\textbf{10933} \\ \textcolor{green}{(-3.87\%)}} \\ \hline

\textbf{EAFD (NTP)} 
& \makecell{\textbf{0.739} \\ \textcolor{green}{(+0.41\%)}} 
& \makecell{\textbf{0.897} \\ \textcolor{green}{(+0.23\%)}} 
& \makecell{\textbf{10671} \\ \textcolor{green}{(-1.55\%)}} \\ \hline

\textbf{EAFD (LLM4ES)} 
& \makecell{\textbf{0.714} \\ \textcolor{green}{(+3.18\%)}} 
& \makecell{\textbf{0.888} \\ \textcolor{green}{(+12.55\%)}} 
& \makecell{\textbf{11108} \\ \textcolor{green}{(-3.09\%)}} \\ \hline
\end{tabular}
\caption{Performance comparison on the Private Dataset across classification and regression tasks. Percentages denote relative change with respect to the corresponding base embedding; for regression, negative values indicate MAE reduction (improvement).}
\label{tab:private_results}
\end{table}

Second, EAFD improves downstream performance when applied to weaker embeddings. Starting from the LLM4ES baseline, which underperforms contrastive encoders, EAFD increases Gender prediction performance by $+12.55\%$ and Age accuracy by $+3.18\%$. These results show that EAFD recovers task-relevant semantic information not captured by the original embedding, raising LLM4ES performance to a level comparable with strong contrastive baselines.

\begin{figure}[h]
    \centering
    \includegraphics[width=0.95\linewidth]{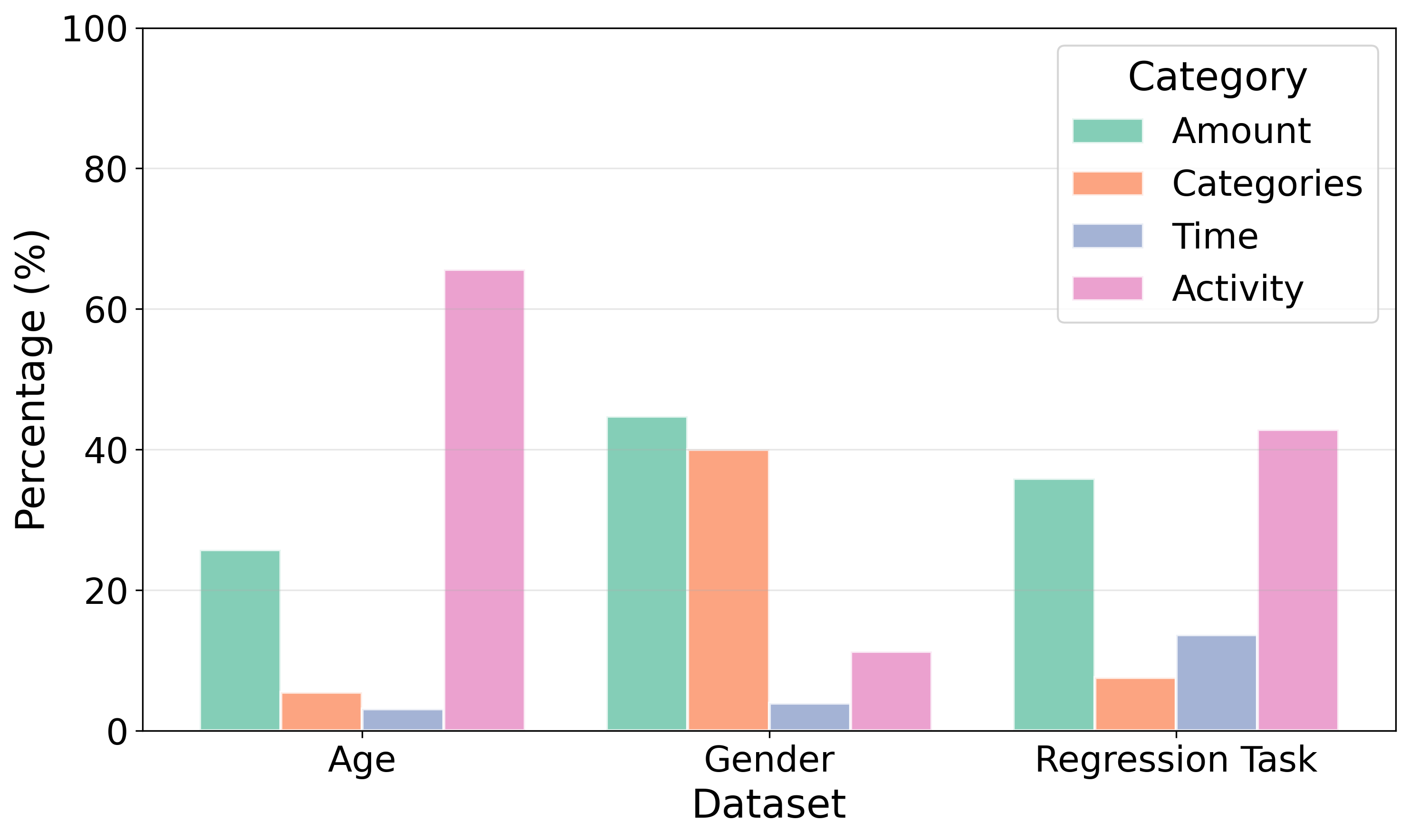}
    \caption{\textbf{Task-adaptive distribution of discovered feature types.} Feature category proportions generated by EAFD for age, gender, and regression tasks.}
    \label{fig:feature_dist}
\end{figure}

Beyond quantitative gains, EAFD demonstrates intrinsic adaptability by tailoring feature generation to the semantic requirements of each target, as visualized in Figure \ref{fig:feature_dist}. For Age prediction, the generator prioritizes Activity patterns ($\approx65\%$), leveraging lifestyle signals encoded in merchant interactions. In contrast, Gender prediction shifts focus to a balanced combination of Amount ($\approx 45\%$) and Categories ($\approx 40\%$), capturing distinct spending magnitudes and preferences. The Regression task triggers a strategic increase in Time-based features, confirming that capturing temporal dynamics is essential for forecasting.

\subsection{Interpretability Analysis}
\label{sec:interpret}

To analyze the semantic content of embeddings, we recovered 43 EAFD features aggregated into four groups: \textit{Amount}, \textit{Categories}, \textit{Time}, and \textit{Activity}. Figure~\ref{fig} reports the alignment score ($R^2$) for three architectures:
\begin{itemize}
\item \textbf{CoLES:} Prioritizes static information. It dominates in \textit{Categories} ($0.816$) and \textit{Activity} ($0.897$) but fails to capture temporal dynamics ($0.327$).
\item \textbf{NTP:} Shows a specialized profile with peak performance in \textit{Activity} ($0.942$), but struggles with explicit features like \textit{Amount} ($0.330$) and \textit{Categories} ($0.272$).
\item \textbf{LLM4ES:} Excels in temporal modeling, surpassing baselines in \textit{Time} ($0.524$). It maintains balanced semantic recovery but retrieves fewer \textit{Activity} markers ($0.599$) than other encoders.
\end{itemize}

Building on this analysis, we next showcase two concrete examples of how the guidance provided by EAFD can be applied in practice.

\subsubsection{Encoder Refinement through Gap Analysis} 
\label{par:refinement}

Standard pretrained encoders, such as CoLES or NTP, often overlook domain-specific patterns because they are optimized using general-purpose self-supervised objectives. EAFD identifies information gaps by detecting features that are insufficiently encoded in the latent space and uses this signal to guide representation refinement.

The analysis of the CoLES encoder, illustrated in  Figure \ref{fig}, highlights a critical deficiency in capturing both numerical distribution patterns (amounts) and temporal dynamics. To mitigate these "blind spots", we enhanced the encoder's input layer with several transformations, including $\log(x)$, $\exp(x)$, and Piecewise Linear Encoding~\cite{gorishniy2022embeddings} for transaction amounts, alongside Time2Vec embeddings~\cite{kazemi2019time2vec} for temporal features.

\begin{table}[h]
\centering
\begin{tabularx}{\linewidth}{|l|p{1.3cm}|X|} 
\hline
\textbf{Category} & \textbf{Base Encoder} & \textbf{Enhanced Encoder} \\
\hline
Amount ($R^2$) & 0.451 & 0.484 \textcolor{green}{(+7.39\%)} \\
Category ($R^2$) & 0.816 & 0.687 \textcolor{red}{(-15.85\%)} \\
Time ($R^2$) & 0.327 & 0.608 \textcolor{green}{(+85.68\%)} \\
Other features ($R^2$) & 0.897 & 0.870 \textcolor{red}{(-3.07\%)} \\
\hline
Downstream (AUC) & 0.835 & 0.845 \textcolor{green}{(+1.20\%)} \\
\hline
\end{tabularx}
\caption{Reconstruction Score $R^2$ of feature classes from embedding representations and downstream performance on the Rosbank dataset between Base and Enhanced Encoder.}
\label{tab:encoder_inprovment}
\end{table}

Table~\ref{tab:encoder_inprovment} compares the reconstruction quality across feature categories and the downstream performances for the enhanced and baseline CoLES encoders. The refined encoder substantially improves over the base model, increasing reconstruction quality for temporal patterns by 85.68\% and improving the modeling of transaction amounts, while reducing MCC precision (-15.85\%). This reallocation of representational capacity toward continuous features results in higher downstream performance, improving churn prediction from $0.835$ to $0.845$ (+1.20\%).






\subsubsection{Privacy-Preserving Feature Erasure} 
\label{par:privacy}

\begin{figure}[h]
    \centering
    \includegraphics[width=1.0\linewidth]{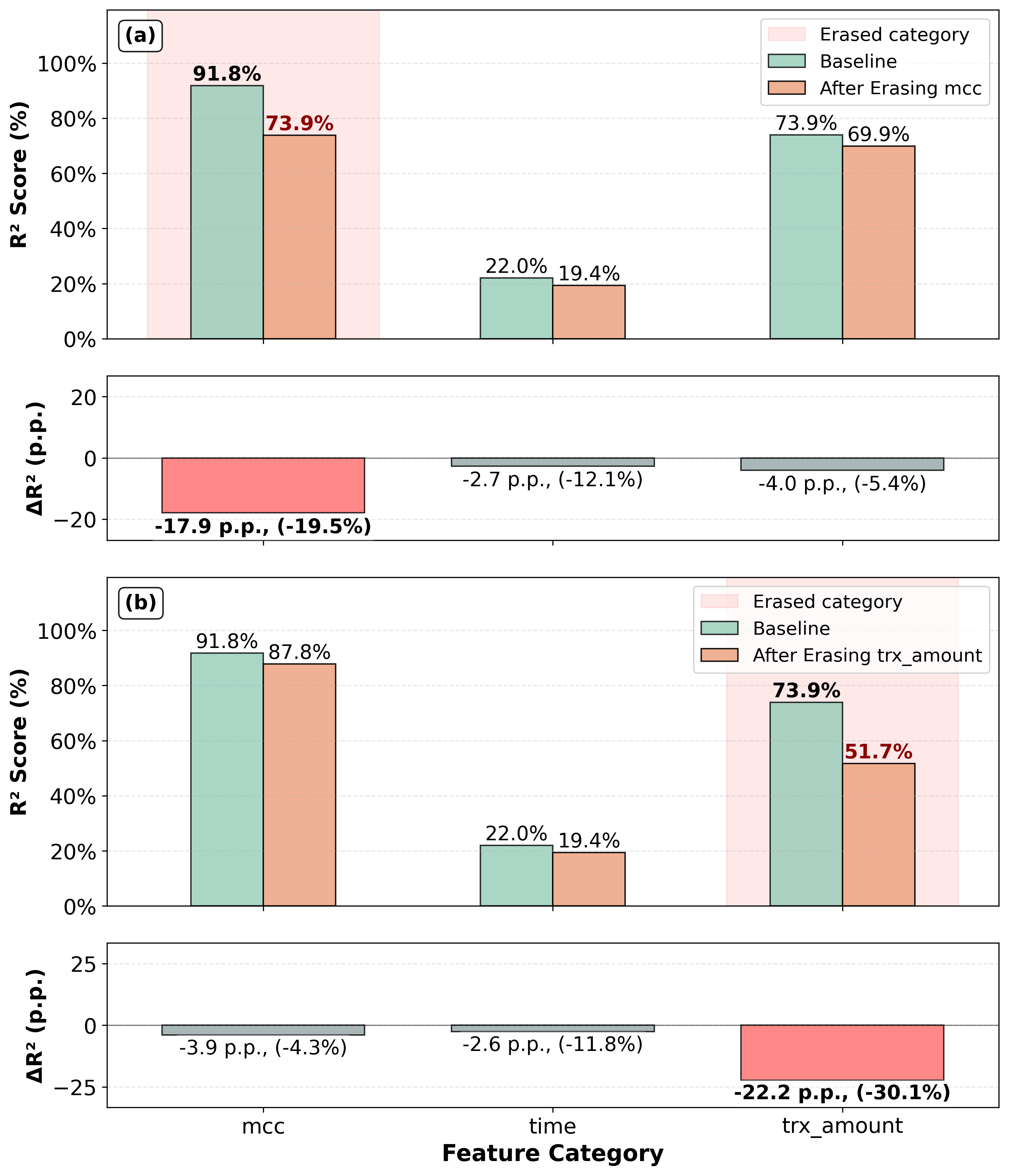}
    \caption{\textbf{Selective feature erasure in embeddings.} Top panels show  $R^2$ reconstruction performance before and after erasing \textit{mcc} (a) and \textit{trx\_amount} (b) feature categories, while the bottom panels report the corresponding relative performance change ($\Delta R^2$).}

    \label{fig:privacy}
\end{figure}

Privacy-preserving feature erasure studies how specific sensitive attributes can be selectively removed from learned embeddings while preserving their utility for downstream tasks. Here, the erasure process is guided by EAFD, which identifies and targets embedding-aligned feature groups for controlled information removal.

We evaluate the effect of removing sensitive information from embeddings using an auxiliary de-correlation regularizer based on the Hilbert–Schmidt Independence Criterion (HSIC). Given embeddings $z \in \mathbb{R}^d$ and a sensitive attribute $s$, the training objective becomes
\begin{equation}
\mathcal{L} = \mathcal{L}_{\text{CoLES}} + \lambda  \mathrm{HSIC}(z, s),
\end{equation}
where $\mathrm{HSIC}(z, s)$ penalizes statistical dependence between the learned representation and the sensitive attribute, and $\lambda$ controls the strength of privacy enforcement.

Figure \ref{fig:privacy} illustrates the obtained results on the DataFusion dataset using CoLES embeddings. We report the predictability ($R^2$) of different feature groups from the embedding before and after applying HSIC-based de-correlation, together with the corresponding change $\Delta R^2$.

Erasing \textit{mcc} features substantially reduces their recoverability ($\Delta R^2 = -17.9$ p.p.) with minimal impact on other attributes. Similarly, erasing \textit{trx\_amount} primarily suppresses the targeted information ($\Delta R^2 = -22.2$ p.p.) while leaving unrelated features largely unaffected. Compared to the baseline embeddings trained without attribute erasing, the downstream ROC-AUC decreases by approximately $1$-$2\%$, depending on which attribute group is erased.

These results demonstrate that the proposed HSIC regularization enables fine-grained, attribute-specific information removal. The effect remains mostly confined to the erased feature group, showing that sensitive information can be selectively reduced with minimal impact on the overall representational capacity of the embedding.

\section{Ablation studies}

In this section, we study which components of the framework are necessary to obtain the reported gains.

\subsection{Impact of LLM Generator Backbone}
\label{sec:llm_selection}

\begin{table}[h]
\centering
\small
\setlength{\tabcolsep}{6pt}
\begin{tabular}{cccc}
\toprule
Llama3.1-8B & Llama3.3-70B & gpt-oss-20B & \textbf{gpt-oss-120B} \\
\midrule
0.835 & 0.856 & 0.859 & \textbf{0.872} \\
\bottomrule
\end{tabular}
\caption{EAFD performance on the Rosbank (AUC) for different LLM generator backbones.}
\label{tab:llm_comparison}
\end{table}

In this ablation we analyze the effect of the LLM backbone used by the feature generator. Table~\ref{tab:llm_comparison} reports downstream performance on the Rosbank dataset across four generator models.

While all models can perform reflective feature generation, reasoning-oriented backbones (gpt-oss) consistently produce more diverse and executable features. Smaller instruction-tuned models more frequently fail during code generation, triggering repeated self-correction and limiting scalability when generating more than 10 features.

\subsection{Performance across Embedding Backbones}

\begin{table}[h]
\centering
\small
\renewcommand{\arraystretch}{1.2}
\begin{tabular}{|l|c|c|}
\hline
\textbf{Method} & \textbf{NTP} & \textbf{CoLES} \\
\hline
Embeddings & 0.648 & 0.738 \\
\hline
Featuretools 
& 0.728~({\color{green}+12.3\%}) 
& 0.760~({\color{green}+3.0\%}) \\
\hline
OpenFE 
& 0.730~({\color{green}+12.7\%})
& 0.753~({\color{green}+2.0\%}) \\
\hline
LightAutoML 
& 0.731~({\color{green}+12.8\%}) 
& 0.760~({\color{green}+3.0\%}) \\
\hline
LLMFE 
& 0.728~({\color{green}+12.3\%})
& 0.760~({\color{green}+3.0\%}) \\
\hline
CAAFE 
& 0.750~({\color{green}+15.7\%}) 
& 0.762~({\color{green}+3.3\%}) \\
\hline
\textbf{EAFD} 
& \textbf{0.773~({\color{green}+19.3\%})} 
& \textbf{0.781~({\color{green}+5.8\%})} \\
\hline
\end{tabular}
\caption{Comparison of automated feature engineering methods across different embeddings. }
\label{tab:tab_comparison}
\end{table}

Table~\ref{tab:tab_comparison} evaluates feature engineering methods across two event-sequence embedding backbones of different quality. EAFD consistently achieves the largest relative improvements on both the weaker NTP embedding ($+19.3\%$) and the stronger CoLES embedding ($+5.8\%$), outperforming all tabular feature engineering baselines in both cases.

Overall, existing LLM-based feature engineering methods are primarily designed for static tabular inputs and do not explicitly model event-sequence structure. When applied on top of embeddings, feature generation operates in an abstract representation space without grounding in the original temporal attributes, which limits the ability to recover complementary sequence-level information. This structural mismatch results in systematically weaker and less stable performance gains compared to EAFD.

\section{Conclusion}

We introduced Embedding-Aware Feature Discovery (EAFD), a principled framework for jointly reasoning over learned embeddings and structured features in event-sequence data. Across four open-source financial benchmarks, EAFD consistently outperformed embedding-only and feature-based baselines, achieving relative gains of up to 5.8\% for state-of-the-art embeddings and up to 19\% for weaker representations. Beyond performance, EAFD diagnosed representational gaps and systematic biases in embedding models and leveraged these signals to design an alternative, more effective training scheme for CoLES, resulting in measurable downstream improvements. By enabling joint feature discovery in the combined embedding–feature space, EAFD produced representations that transferred effectively across multiple classification and regression targets on large-scale industrial dataset.

Looking forward, this work can be extended toward multi-agent orchestration, where feature generation, selection, and refinement are handled by specialized agents coordinated via embedding-aware signals. Embedding-aware discovery can also be integrated with ensemble embeddings, enabling joint reasoning over multiple representations and adaptive feature exploration under representational uncertainty. More broadly, EAFD's interpretation signals provide a foundation for automatic encoder and model tuning, where identified information gaps guide architecture, sampling, and training objectives, closing the loop between representation learning and system optimization.

\bibliographystyle{named}
\bibliography{ijcai26}

\end{document}